\documentclass[runningheads]{llncs}
\usepackage[T1]{fontenc}
\usepackage[a4paper, 
    left=3.17cm, 
    right=3.17cm, 
    top=2.54cm, 
    bottom=2.54cm]{geometry}
\usepackage{float}
\usepackage{listings} 
\usepackage{xcolor}
\usepackage{subcaption}
\usepackage{graphicx}
\usepackage{siunitx}
\usepackage{tabularx} 
\usepackage{booktabs} 
\usepackage{colortbl} 
\usepackage{amsmath} 
\usepackage{multicol} 
\usepackage{multirow} 
\usepackage{microtype} 
\begin{document}
\title{IFDNS: An Iterative Feedback-Driven Neuro-Symbolic Method for Faithful Logical Reasoning}
\titlerunning{IFDNS}
%
\author{Xiaoheng Wang\inst{1}\orcidID{0009-0005-3335-2322} \and
Tongxuan Liu\inst{1}\orcidID{0009-0007-2634-2788} \and
Zi Gong\inst{1}\orcidID{0009-0004-8863-3677} \and
Xianzhe Dong\inst{1}\orcidID{0009-0007-6943-9432} \and
Yuting Zeng\inst{1}\orcidID{0009-0006-1719-5424} \and
Minhan Hu\inst{1}\orcidID{0009-0007-3995-8821} \and
Weizhe Huang\inst{1}\orcidID{0009-0005-0320-6068} \and
Jing Li\inst{1}\orcidID{0000-0001-6761-7687}\thanks{Corresponding author}
}

\authorrunning{X. Wang et al.}
%
\institute{College of Computer Science and Technology, University of Science and Technology of China, Hefei, Anhui Province, China\\
\email{\{wangxiaoheng, tongxuan.ltx, gong513, xianzhedong, yuting\_zeng, hmh, hwz871982879\}@mail.ustc.edu.cn}\\
\email{lj@ustc.edu.cn}}

\maketitle               
\begin{abstract}

Large language models (LLMs) have demonstrated impressive capabilities across a wide range of reasoning tasks, including logical and mathematical problem-solving. While prompt-based methods like Chain-of-Thought (CoT) can enhance LLM reasoning abilities to some extent, they often suffer from a lack of faithfulness, where the derived conclusions may not align with the generated reasoning chain. To address this issue, researchers have explored neuro-symbolic approaches to bolster LLM logical reasoning capabilities. However, existing neuro-symbolic methods still face challenges with information loss during the process. To overcome these limitations, we introduce Iterative Feedback-Driven Neuro-Symbolic (IFDNS), a novel prompt-based method that employs a multi-round feedback mechanism to address LLM limitations in handling complex logical relationships. IFDNS utilizes iterative feedback during the logic extraction phase to accurately extract causal relationship statements and translate them into propositional and logical implication expressions, effectively mitigating information loss issues. Furthermore, IFDNS is orthogonal to existing prompt methods, allowing for seamless integration with various prompting approaches. Empirical evaluations across six datasets demonstrate the effectiveness of IFDNS in significantly improving the performance of CoT and Chain-of-Thought with Self-Consistency (CoT-SC). Specifically, IFDNS achieves a +9.40\% accuracy boost for CoT on the LogiQA dataset and a +11.70\% improvement for CoT-SC on the PrOntoQA dataset.  

\keywords{Logical Reasoning  \and Large Language Model \and Reasoning.}
\end{abstract}

\section{Introduction}
Large language models (LLMs) such as GPT-4o \cite{achiam2023gpt}, o1 \cite{zhong2024evaluation}, DeepSeek-R1 \cite{guo2025deepseek}, and Gemma \cite{team2024gemma} have demonstrated exceptional capabilities across various NLP tasks. However, studies \cite{arkoudas2023gpt,liu2023evaluating,zhong2024evaluation,jahin2025unveiling} reveal that significant challenges persist in complex logical reasoning and mathematical problem-solving scenarios. Researchers have proposed various methods to enhance LLMs' reasoning abilities, such as Chain-of-Thought (CoT) \cite{kojima2022large,wei2023chainofthought,nye2021show}, Chain-of-Thought with Self-Consistency (CoT-SC) \cite{wang2022self}, Tree-of-Thought (ToT) \cite{yao2024tree}, and STaR \cite{eric2022star}, which improve LLMs' performance through prompt-based approaches. However, these methods occasionally exhibit unfaithful reasoning \cite{bao2024llms,lanham2023measuring,lyu2023faithful,turpin2024language}, leading to conclusions that contradict previously generated reasoning chains. To address unfaithful reasoning, approaches like Faithful Chain-of-Thought \cite{lyu2023faithful}, LINC \cite{olausson2023linc}, Logic-LM \cite{pan2023logic}, and SatLM \cite{ye2024satlm} integrate LLMs with symbolic reasoning methods. These frameworks transform logical problems into formal expressions, utilize external symbolic solvers to derive symbolic results, and then interpret these results via LLMs or specialized interpreters. Nevertheless, Logic-of-Thought (LoT) \cite{liu2024logic} identifies that such methods suffer from information loss during logical expression extraction, resulting in incorrect intermediate reasoning steps and outcomes.

To address the information loss issue during logical expression extraction, \cite{liu2024logic} proposes supplementing the original prompt with descriptions of generated logical propositions. This approach preserves the information from the original prompt while incorporating additional logical information derived through a neuro-symbolic method. However, although this method mitigates information loss via supplementary augmentation, the inherent hallucination issues of LLMs may still occasionally cause failures in the logical extraction phase, such as redundant expressions, omissions of logical relationships, and deviations between logical propositions and formal expressions \cite{liu2024logic}.

\begin{figure}[t]
    \centering
    \includegraphics[width=1\linewidth]{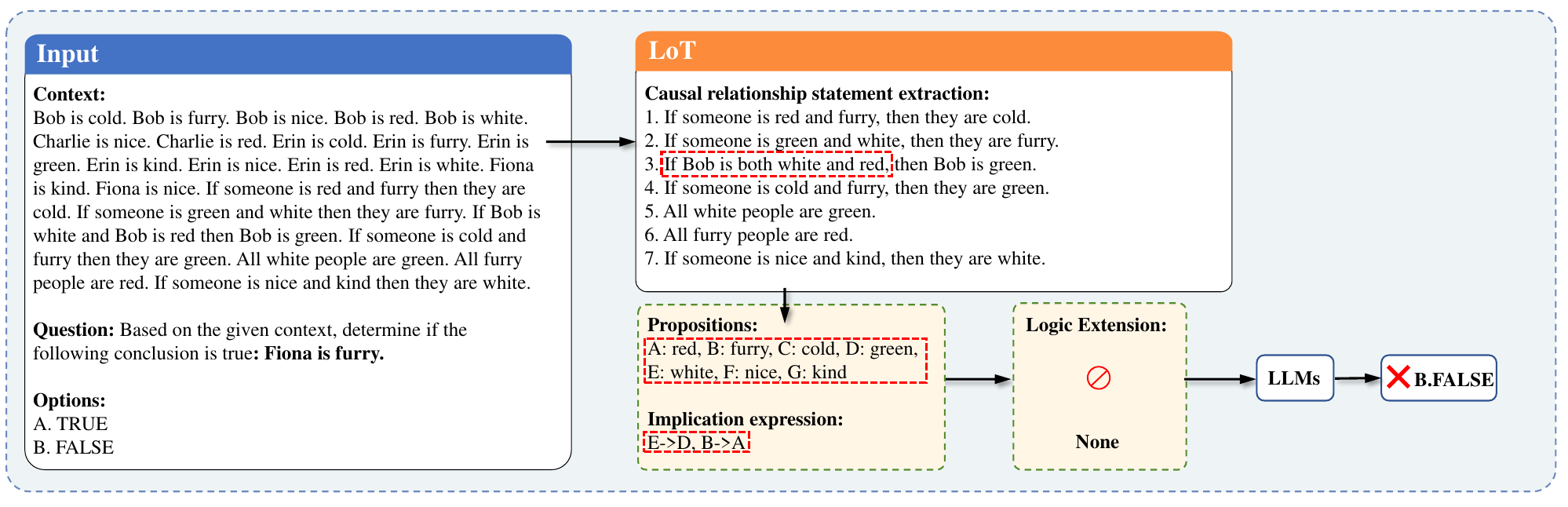}
    \caption{In the LoT case from the ProofWriter dataset, LoT firstly fails to comprehensively and accurately extract causal relationship statements. Secondly, it lacks the capability to fully interpret all propositions within the context and translate them into corresponding implication expressions, thereby preventing logical extension.}
    \label{fig:fig1}
\end{figure}

As illustrated in Figure \ref{fig:fig1}, we find that failures in the logical extraction phase primarily stem from LLMs' inability to comprehensively understand all propositions and implication expressions contained in the context in a single pass. The LoT method converts an input multiple-choice question-answering task (MCQA) into simple logical propositions and implication expressions via an LLM, followed by logical deduction of the implication expressions. However, when extracting causal statements from the context, LoT produces ambiguous statements such as "If Bob is both white and red." which hinders the LLM from later extracting them into two distinct propositions: "Bob is white." and "Bob is red." During the subsequent proposition and implication expression extraction process, LoT can only vaguely identify simple propositions in the context, such as "A: red.", "B: furry." and "C: cold." These propositions lack subject and quantifier information, leading to erroneous merging of objects across different subject scenarios (e.g., symbolizing both "Bob is white." and "All white people." as "E: white."). This failure to distinguish entity ownership relationships or handle complex logical conjunctions results in a coarse-grained symbolization process that discards critical logical information. Consequently, the subsequent logical extension phase cannot derive implicit multi-layered relationships, ultimately causing errors in LLMs' complex multi-hop reasoning.

Motivated by this observation, we propose IFDNS, a novel few-shot prompting method that addresses the limitations of LLMs in processing intricate logical relationships through a multi-round feedback mechanism. Specifically, IFDNS employs the LLM to extract causal statements and convert them into propositions and logical implication expressions, applying multi-round feedback at both stages. During the logical extension phase, Python-implemented logical rules extend the logical expressions. In the logical translation phase, the extended expressions are translated by the LLM into natural language descriptions and integrated into the context. Finally, in the word order reordering phase, the contextual content is reordered to form the final input prompt.

To validate the effectiveness of IFDNS, we conducted extensive experiments evaluating its capabilities across different models, multiple prompting methods, and diverse datasets. The results demonstrate that IFDNS substantially enhances the performance of various prompting methods, including Direct, CoT, and CoT-SC. Specifically, IFDNS significantly improved the accuracy of the Direct method on the PrOntoQA dataset by +13.90\%. Furthermore, it boosted the accuracy of CoT on the LogiQA dataset by +9.40\%, while also achieving an +11.70\% accuracy improvement for CoT-SC on the PrOntoQA dataset.

The main contributions of this work are as follows: 
\begin{itemize}
\item[1.] We propose IFDNS, a novel prompting method that addresses the information loss issue in logical extraction phases of existing neuro-symbolic approaches through a multi-round feedback mechanism.
\item[2.] Leveraging the orthogonal capability of IFDNS, we integrate it with various prompting methods such as CoT and CoT-SC.
\item[3.] We validate the effectiveness of IFDNS across different models and logical reasoning tasks, demonstrating its ability to broadly enhance the performance of diverse prompting methods.
\end{itemize}

\section{Preliminary}
\subsection{Multiple-choice Question-answering Task}
Our study focuses on MCQA tasks. A QA task is formally defined as a tuple $(Q, C, O, A)$, where $Q$ denotes the question, $C$ represents the context providing background information, $O = (o_1, o_2, \ldots, o_k)$ is the set of answer options, and $A$ is the ground-truth answer. The objective of LLMs is to accurately answer $Q$ by generating a sequence of tokens $T = (t_1, t_2, \ldots, t_m)$ through a sequence of logically coherent intermediate reasoning steps $RP = (r_{p_1}, r_{p_2}, \ldots, r_{p_m})$, where each step $r_{p_i}$ corresponds to a subset of tokens $(t_{l_i}, t_{r_i}) \subseteq T$.

\subsection{Introduction to Propositional Logic}
Propositional logic studies logical relationships between propositions, focusing on \textit{propositions} (statements with determinate truth values, e.g., "It is raining"). We denote specific propositions using uppercase letters (e.g., $A$, $B$, $C$) and arbitrary propositions using lowercase letters (e.g., $p$, $q$, $r$). Propositions are classified as atomic (indivisible statements) or compound (formed via logical connectives). This paper primarily uses three fundamental connectives:
\begin{itemize}
    \item \textbf{Negation} $\lnot$ (not): $\lnot p$ is true $iff$ $p$ is false.
    \item \textbf{Conjunction} $\land$ (and): $p \land q$ is true $iff$ both $p$ and $q$ are true.
    \item \textbf{Implication} $\rightarrow$ (if-then): $p \rightarrow q$ is false only when $p$ is true and $q$ is false.
\end{itemize}

Propositional formulas are recursively defined using atomic propositions and connectives. Their truth values are determined via truth assignments, yielding three categories: tautologies (always true, e.g., $p \rightarrow p$), contradictions (always false, e.g., $\lnot(p \rightarrow \lnot p)$), and satisfiable formulas (true under some assignments). Logical inference rules define derivational relationships between formulas. We use $\Rightarrow$ for entailment (one formula deriving another) and $\Leftrightarrow$ for logical equivalence. Key axioms in our system include:
\begin{enumerate}
    \item \textbf{Double Negation Law}: $\lnot \lnot p \Leftrightarrow p$
    \item \textbf{Transitive Law}: $(p \rightarrow q) \land (q \rightarrow r) \Rightarrow (p \rightarrow r)$
    \item \textbf{Contraposition Law}: $p \rightarrow q \Leftrightarrow \lnot q \rightarrow \lnot p$
\end{enumerate}

The formula can be transformed into an implication expression\textbf{ }(such as $p \land q \Leftrightarrow \lnot (p\rightarrow \lnot q)$) through equivalent transformation. Propositional logic is the foundation of computer circuit design and program conditional reasoning \cite{logic-driven,buning1999propositional}. The logical system proposed here is relatively simple. In Section \ref{sec:mintishibie}, we will further explain how our work is based on common propositional logic rules and partially extended to strengthen the extraction of subjects and quantifiers in propositions.

\section{Methodology}\label{sec:mintishibie}
\begin{figure}[t]
    \centering
    \includegraphics[width=1\linewidth]{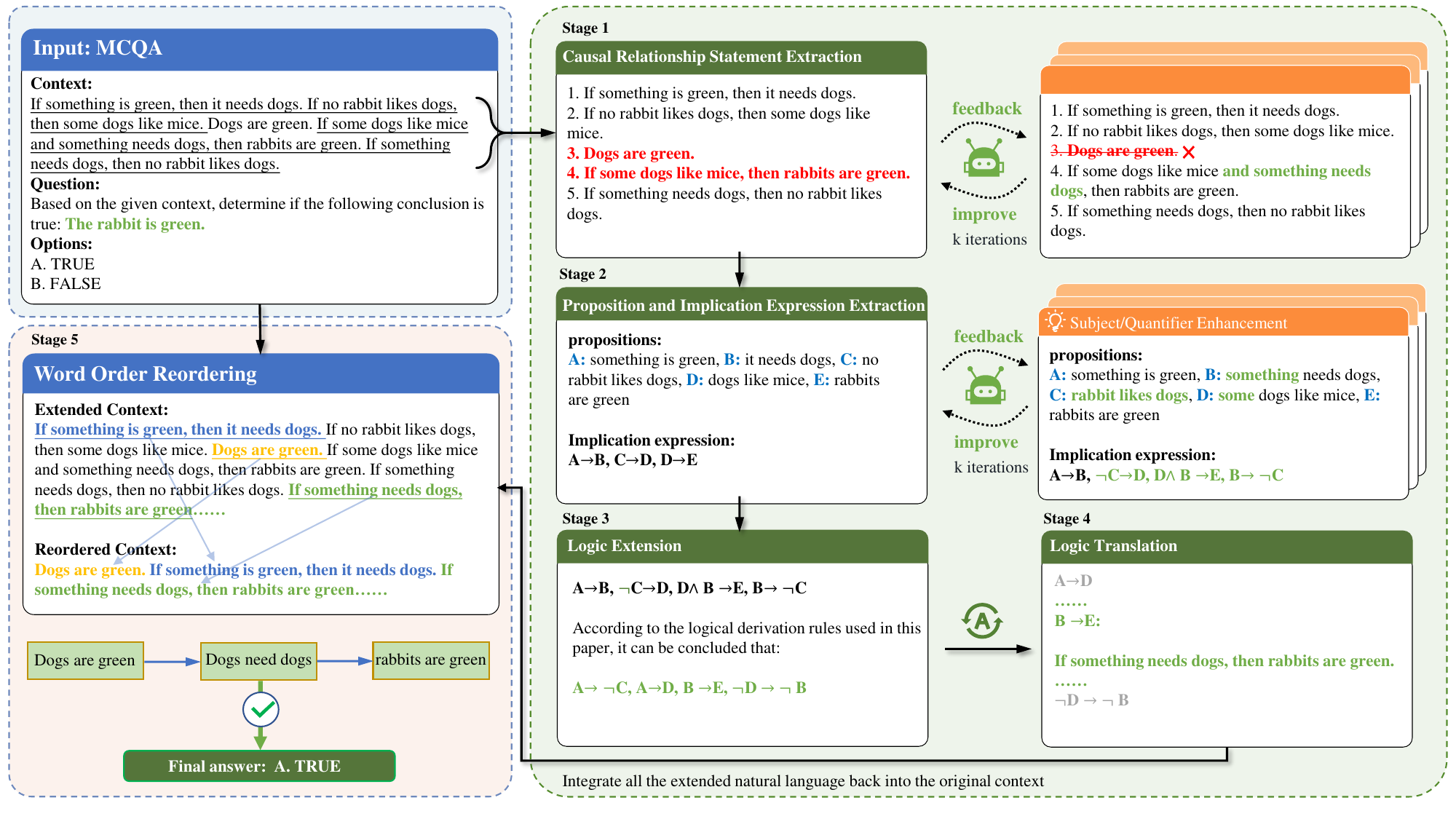}
    \caption{The IFDNS framework operates through five sequential stages: (1) Upon receiving MCQA input, LLMs extract causal relationship statements from the context; (2) Extract propositions and implication expressions from causal relationship statements, while conducting multiple rounds of feedback for both the first and second stages. The statements highlighted in red in the diagram are those that are about to be modified or removed after the feedback process; (3) Python-based deductive engine extends implications via logical inference laws; (4) Translation of extended implications into natural language for context augmentation; (5) Word order reordering module optimizes context sequencing to enable LLM-driven end-to-end reasoning.}
    \label{fig:enter-label}
\end{figure}

\textbf{Overview. }  Figure \ref{fig:enter-label} illustrates the overall framework of IFDNS, which comprises five stages. The first stage is causal relationship statement extraction, where LLMs are employed to extract causal statements from the context. Subsequently, the extracted causal statements are transformed into propositions and logical implication expressions. Next, during the logical extension stage, a Python-implemented logical reasoning program extends these logical expressions. In the logical translation stage, the extended logical expressions are translated back into natural language via LLMs and integrated into the context. Finally, in the word order reordering stage, the original contextual content undergoes systematic word order rearrangement to formulate the ultimate input prompt. The following sections provide detailed explanations of these five stages.

\textbf{Causal Relationship Statement Extraction.}
In this stage, we prompt LLMs to generate a set of causal sentences $\mathcal{C}$. These sentences are output according to their original semantic order in the context. Multi-round feedback is then applied to $\mathcal{C}$, leveraging few-shot guidance to prompt the LLM to evaluate the extracted causal statements based on five metrics: informational completeness, faithfulness to original intent, logical consistency, evidential relevance, and clarity of expression. Each metric is scored from 1 to 5. Based on the feedback from each iteration, the LLM is driven to generate improved causal statements, iterating this process $k$ times.

\textbf{Proposition and Implication Expression Extraction. } 
LLMs are utilized to refine a set of propositional symbols $\mathcal{P}$ and logical implication expressions $\mathcal{E}$ from the provided collection $\mathcal{C}$. In this step, LLMs identify semantically similar propositions and unify them under the same propositional symbol. They also parse logical relationships between propositions and process them according to their natural language expressions. For propositions expressing negation, the negation operator $\lnot$ is appended. If a causal relationship exists between two propositions, the implication symbol $\rightarrow$ is used to connect them. For example, LLMs might extract the semantically equivalent description "someone is able to use a computer" from two distinct sentences, symbolizing it as proposition $B$. By analyzing logical relationships, the LLM may apply $\lnot$ to $B$ and another proposition $A$, generating an implication expression $\lnot A \rightarrow \lnot B$. To address insufficient inductive generalization during symbol extraction, we enhance proposition recognition accuracy through refined symbol processing: statements with different subjects and quantifiers are assigned distinct propositional symbols for fine-grained extraction. For instance, given the causal statement: 
\begin{enumerate}
    \item If something chases the bald eagle and has a dependence on the lion, then it must also have a dependence on the mouse.  
    \item If the lion chases the bald eagle and notices the dog, then it also observes the mouse.
\end{enumerate}

We obtain:
\textbf{Propositions:} A: something chases the bald eagle, B: something has a dependence on the lion, C: something has a dependence on the mouse, D: the lion chases the bald eagle, E: the lion notices the dog, F: the lion observes the mouse.

\textbf{Implication Expressions:} $A\land B\rightarrow C, D\land E\rightarrow F$.

By strengthening the extraction mechanisms for subjects and quantifiers, we simultaneously approximate the inductive capacity of first-order logic while avoiding frequent symbol extraction errors that occur when directly generating first-order logic expressions with LLMs. In this process, we also implement a multi-round feedback mechanism for both $\mathcal{P}$ and $\mathcal{E}$: leveraging few-shot guidance to prompt the LLM to evaluate $\mathcal{P}$ and $\mathcal{E}$, and based on the feedback from each iteration, driving the LLM to generate optimized versions of $\mathcal{P}$ and $\mathcal{E}$. This procedure is repeated for $k$ iterations.

\textbf{Logic Extension.}  
During the logic extension stage, we apply laws of logical inference to the implication expression set $\mathcal{E}$ obtained from the previous stage. We developed a Python program implementing propositional logic-based deduction for implication expressions. For basic demonstration, consider the extracted implication expressions $\lnot A \rightarrow \lnot B$ and $\lnot B \rightarrow \lnot C$ as input to our logical deduction program. Through logic extension based on the law of transitivity and contraposition  we ultimately derive the new expression $C \rightarrow A$, which will be utilized in the next stage. IFDNS also supports deduction of implication relationships involving complex conjunctions. For example, given implication expressions $A \land B \rightarrow C,\ A \rightarrow B$, we derive the extended implication $A \rightarrow C$, given implication expressions $A \land B \land C \rightarrow D,\ A \rightarrow B,\ B \rightarrow C$, we derive extended implications $A \rightarrow C,\ A \rightarrow D$.

\textbf{Logical Translation.}  
In the logical translation stage, we employ LLMs to convert the generated extended implication expressions into natural language, thereby forming new components of the original input prompt. Through this approach, we integrate the derived logical information into the original prompt, thereby uncovering the underlying logical information embedded in the text.

\textbf{Word Order Reordering.}  
Existing studies \cite{prontoqa,premise_order} have demonstrated that in mathematical problem-solving and logical reasoning domains, varying input premise orders can significantly affect the reasoning outcomes of LLMs. Although neuro-symbolic methods mitigate this issue by converting model inputs into logical symbols, our approach reveals that even after integrating derived logical information into original prompts, premise order randomness persists. To address this, we extract original contexts from the R-GSM dataset \cite{premise_order} and ProofWriter dataset \cite{ruletaker}, manually rearrange their word order to create stepwise deduction-friendly sequences, and employ few-shot prompting to guide LLMs in autonomously performing word order reorganization on the logically extended contextual inputs.

\section{Experiments}
\subsection{Tasks and Datasets}
In our experiments, we utilized six logical reasoning datasets and one mathematical reasoning dataset: ReClor~\cite{yu2020reclor}, LogiQA~\cite{liu2020logiqa}, RuleTaker~\cite{ruletaker}, ProofWriter~\cite{ruletaker}, FOLIO~\cite{han2022folio}, PrOntoQA~\cite{prontoqa}, and R-GSM~\cite{premise_order}.

\begin{itemize}
\item \textbf{LogiQA}: Designed to assess human logical reasoning capabilities, this dataset is constructed from logical reasoning questions in publicly available national civil service exams. Each entry contains a context, a question, and four answer options with one correct choice.
\item \textbf{ReClor}: A reading comprehension dataset requiring logical reasoning, comprising questions extracted from standardized tests (e.g., LSAT and GMAT). Each instance includes a context passage, a question, and four answer options with a single correct answer.
\item \textbf{RuleTaker}: Programmatically generated using logical connectives ($\land$, $\lnot$, $\rightarrow$). Each problem consists of a context and a conclusion requiring verification.
\item \textbf{ProofWriter}: Focused on formal logical reasoning, this dataset evaluates multi-step proof generation from given rules and facts. It assesses model performance on complex reasoning tasks through automatically generated problems.
\item \textbf{PrOntoQA}: A synthetic QA dataset where each example is generated from first-order logic representations of synthetic world models. Each question is accompanied by detailed reasoning chains specifying the steps to derive correct answers.
\item \textbf{FOLIO}: Specialized for first-order logic reasoning, this dataset evaluates models' capacity to understand and manipulate first-order logical expressions.
\item \textbf{R-GSM}: A mathematical reasoning dataset based on GSM8K, designed to test LLMs' sensitivity to premise order through systematic sentence rearrangement in problem statements.
\end{itemize}

\subsection{Baselines}
We consider three prompting methods and two neurosymbolic approaches. The prompting methods include: (1) \textbf{Direct}: Directly answering dataset questions using LLMs without intermediate reasoning; (2) \textbf{CoT} \cite{zero-shot,cot,nye2021show}: Using chain-of-thought prompting to require LLMs to perform step-by-step reasoning before answering; (3) \textbf{CoT-SC} \cite{con-sc}: Combining chain-of-thought reasoning with majority voting, where LLMs generate multiple answer trajectories and select the most consistent response. The neurosymbolic approaches include: (4) \textbf{LoT} \cite{liu2024logic}: Generating augmented logical information from input contexts using propositional logic and integrating derived logical expressions as additional prompts; (5) \textbf{Logic-LM} \cite{panlogiclm23}: Transforming natural language problems into symbolic formulae via LLMs followed by verifiable reasoning using symbolic solvers, with a self-refinement module that leverages solver error messages to revise formalizations. We evaluate IFDNS against these five baselines across all six datasets.

    \subsection{Experiment Setup}
Theoretically, IFDNS imposes no restrictions on the types of LLMs employed. For our experiments, we utilize three foundational models to test the upper bounds of LLM-based logical reasoning: (1) \texttt{GPT-4o} (\texttt{GPT-4o-2024-05-13}) and \texttt{DeepSeek-V3} as massively parameterized LLMs, and (2) the more cost-efficient \texttt{GPT-4o-mini} (\texttt{GPT-4o-mini-2024-07-18}). Default hyperparameters are set as: temperature=0.3 with top\text{-}k=1 for base inference. For CoT-SC, we configure temperature=1.0 (with $n$=5 reasoning paths) and set multi-round feedback iterations $k$=3.

\textbf{Main experiments.} Using IFDNS, we compare two model variants against five baselines across these six datasets. For ReClor, we randomly sample 200 instances from its validation set to specifically evaluate models' deductive reasoning capabilities. Since LogiQA was originally adapted from Chinese civil service exam questions and the English version\footnote{LogiQA dataset: https://github.com/lgw863/logiqa-dataset} provided by the original authors contains numerous errors, we employ \texttt{DeepSeek-V3} to retranslate the Chinese dataset followed by manual correction, subsequently selecting 200 curated samples. Regarding RuleTaker, we randomly extract 200 test instances with reasoning depths randomly distributed across 1-5 levels. For ProofWriter, we select 200 validation samples with maximum hop count of 5. In FOLIO, we extract 135 validation instances. For PrOntoQA, we generate 200 samples programmatically using its source code, ensuring 5-hops, false ontology conditions, and randomized word order.
\begin{table}[t]
\caption{The main results of IFDNS integrated with various baseline methods across six reasoning datasets, using models \texttt{GPT-4o} and \texttt{GPT-4o-mini}, are presented as accuracy percentages(\%). \underline{\textbf{Bold and underlined}} values indicate the highest accuracy achieved, with \textcolor{green}{green} highlighting performance improvements and \textcolor{red}{red} denoting declines.}
\centering
\scriptsize
\sisetup{table-format=2.2, 
table-number-alignment=center}
\begin{tabularx}{\textwidth}{c *{6}{X}}
\toprule

\multirow{2}{*}{\textbf{Methods}} &
\multicolumn{6}{c}{\textbf{Datasets}} \\
\cmidrule(lr){2-7}
& {\text{LogiQA}} & {\text{ReClor}} & {\text{RuleTaker}} & {\text{ProofWriter}} & {\text{PrOntoQA}} & {\text{FOLIO}} \\
\rowcolor{gray!20} 
\midrule
\multicolumn{7}{c}{\textbf{GPT-4o-mini}} \\
Direct       & 59.20 & 75.40 & 60.60 & 61.40 & 58.50 & 85.33 \\
Direct+IFDNS   & 60.40{\textcolor{green}{$\uparrow$}\textcolor{green}{1.20}} & 77.50{\textcolor{green}{$\uparrow$}\textcolor{green}{2.10}} & 60.20{\textcolor{red}{$\downarrow$}\textcolor{red}{0.40}} & 65.00{\textcolor{green}{$\uparrow$}\textcolor{green}{3.60}} & 60.50{\textcolor{green}{$\uparrow$}\textcolor{green}{2.00}} & 86.67{\textcolor{green}{$\uparrow$}\textcolor{green}{1.34}}\\
\midrule
CoT          & 60.20 & 75.50 & 62.30 & 68.20 & 69.40 & 86.81 \\
CoT+IFDNS      & 63.70{\textcolor{green}{$\uparrow$}\textcolor{green}{3.50}} & 77.00{\textcolor{green}{$\uparrow$}\textcolor{green}{1.50}} & 68.50{\textcolor{green}{$\uparrow$}\textcolor{green}{6.20}} & 74.90{\textcolor{green}{$\uparrow$}\textcolor{green}{6.70}} & 76.80{\textcolor{green}{$\uparrow$}\textcolor{green}{7.40}} & 89.63{\textcolor{green}{$\uparrow$}\textcolor{green}{2.82}} \\
\midrule
CoT-SC       & 61.70 & 75.80 & 62.80 & 69.50 & 68.90 & 87.26 \\
CoT-SC+IFDNS   & \underline{\textbf{67.90}}{\textcolor{green}{$\uparrow$}\textcolor{green}{6.20}} & \underline{\textbf{78.00}}{\textcolor{green}{$\uparrow$}\textcolor{green}{2.20}} & \underline{\textbf{66.30}}{\textcolor{green}{$\uparrow$}\textcolor{green}{3.50}} & \underline{\textbf{76.60}}{\textcolor{green}{$\uparrow$}\textcolor{green}{7.10}} & \underline{\textbf{80.60}}{\textcolor{green}{$\uparrow$}\textcolor{green}{11.70}} & \underline{\textbf{91.11}}{\textcolor{green}{$\uparrow$}\textcolor{green}{3.85}} \\
\midrule
\rowcolor{gray!20} 
\multicolumn{7}{c}{\textbf{GPT-4o}} \\
Direct       & 62.30 & 83.60 & 61.80 & 67.40 & 65.00 & 85.48 \\
Direct+IFDNS   & 68.20{\textcolor{green}{$\uparrow$}\textcolor{green}{5.90 }} & 88.50{\textcolor{green}{$\uparrow$}\textcolor{green}{4.90 }}
 & 69.50{\textcolor{green}{$\uparrow$}\textcolor{green}{7.70}} & 73.00{\textcolor{green}{$\uparrow$}\textcolor{green}{5.60}}
 & 78.90{\textcolor{green}{$\uparrow$}\textcolor{green}{13.90}}
 & 87.41{\textcolor{green}{$\uparrow$}\textcolor{green}{1.93 }} \\
\midrule
CoT          & 65.70 & 86.90 & 75.20 & 79.90 & 84.50 & 85.63 \\
CoT+IFDNS      & 75.10{\textcolor{green}{$\uparrow$}\textcolor{green}{9.40}}
 & 89.30{\textcolor{green}{$\uparrow$}\textcolor{green}{2.40}}
 & \underline{\textbf{79.00}}{\textcolor{green}{$\uparrow$}\textcolor{green}{3.80}}
 & 83.00{\textcolor{green}{$\uparrow$}\textcolor{green}{3.10}}
 & 92.70{\textcolor{green}{$\uparrow$}\textcolor{green}{8.20}}
 & \underline{\textbf{91.85}}{\textcolor{green}{$\uparrow$}\textcolor{green}{6.22}} \\
\midrule
CoT-SC       & 67.10 & 88.00 & 75.30 & 85.60 & 86.50 & 88.30 \\
CoT-SC+IFDNS   & \underline{\textbf{76.40}}{\textcolor{green}{$\uparrow$}\textcolor{green}{9.30 }} & \underline{\textbf{92.00}}{\textcolor{green}{$\uparrow$}\textcolor{green}{4.00 }}
 & 77.50{\textcolor{green}{$\uparrow$}\textcolor{green}{2.20  }}
 & \underline{\textbf{88.00}}{\textcolor{green}{$\uparrow$}\textcolor{green}{2.40 }}
 & \underline{\textbf{97.80}}{\textcolor{green}{$\uparrow$}\textcolor{green}{11.30}}
 & 88.89{\textcolor{green}{$\uparrow$}\textcolor{green}{0.59}}
 \\
\bottomrule
\end{tabularx}

\label{tab:main_test}
\end{table}

\textbf{CoT+IFDNS vs. CoT+LoT Comparison.} We compare the CoT+IFDNS and CoT+LoT methods across six logical reasoning datasets, investigating the effectiveness differences between IFDNS and LoT as logic-augmented enhancement approaches.

\textbf{CoT+IFDNS vs. Logic-LM Comparison.} We compare CoT+IFDNS with Logic-LM on four logical reasoning datasets. The Logic-LM method falls back to CoT when its generated neurosymbolic representations fail to be parsed by external symbolic solvers. This comparative design between CoT+IFDNS and Logic-LM objectively evaluates the robustness of symbolic parsing in logical enhancement frameworks.

\textbf{Ablation Study. } In addition to the aforementioned baselines, we conduct ablation studies to evaluate the impact of individual modules on the overall framework. All ablation experiments are performed on the \texttt{GPT-4o} with the following variants: (1) \textbf{w/o Multi-round Feedback}: IFDNS without the two-stage symbolic verification (multi-round feedback); (2) \textbf{w/o Subject/Quantifier Enhancement}: Directly extract the coarsest-grained propositions and corresponding symbolic expressions without strengthening the extraction of subjects and quantifiers in propositions; (3) \textbf{w/o Reordering}: Removing the final word order reordering stage, allowing the LLM to reason over randomly ordered data.

\subsection{Main Results}
In this section, we integrate IFDNS with three baseline prompting methods to examine whether their combination enhances reasoning performance across six datasets, as detailed in Table~\ref{tab:main_test}. Overall, IFDNS improves all three prompting methods to varying degrees on both \texttt{GPT-4o-mini} and \texttt{GPT-4o}. Specifically, CoT-SC+IFDNS achieves the highest accuracy on all six datasets with \texttt{GPT-4o-mini} and four datasets with \texttt{GPT-4o}, while CoT+IFDNS attains the highest accuracy (79.00\%) on the RuleTaker dataset under \texttt{GPT-4o}. Across all experimental configurations ($3~\text{baselines} \times 6~\text{datasets} \times 2~\text{models} = 36$ comparisons), IFDNS enhances accuracy in 35 cases. The sole exception occurs with Direct+IFDNS on the \texttt{GPT-4o-mini}-based RuleTaker dataset, showing a marginal 0.40\% decrease compared to the original prompting method. This minor decline remains statistically insignificant given the dataset's limited sample size.

On the most challenging datasets LogiQA and PrOntoQA, we observe that CoT-SC+IFDNS significantly outperforms the original prompting methods with improvements of +6.20\% (LogiQA, \texttt{GPT-4o-mini}), +11.70\% (PrOntoQA, \texttt{GPT-4o-mini}), +9.30\% (LogiQA, \texttt{GPT-4o}), and +11.30\% (PrOntoQA, \texttt{GPT-4o}). This phenomenon suggests that IFDNS combined with other prompting methods achieves maximal performance gains on the most complex logical reasoning tasks. A potential explanation lies in PrOntoQA's fully randomized sentence order, which impairs both human and LLM reasoning capabilities \cite{premise_order}. The word order reordering module in IFDNS effectively addresses this challenge, as demonstrated by ablation studies in Section~\ref{sec:ablation}.

On the RuleTaker dataset where \texttt{GPT-4o} generally exhibits lower accuracy compared to other datasets, CoT+IFDNS  achieves the best performance. Across other datasets, CoT+IFDNS significantly improves the baseline CoT prompting method's accuracy. In 12 comparative evaluations between CoT+IFDNS and CoT-SC, the former outperforms the latter in 11 cases. This demonstrates that CoT+IFDNS, when operationalized as an integrated framework, surpasses the CoT-SC method and establishes itself as a complete reasoning system with strong logical capabilities.

\begin{table}[t]
\caption{Performance comparison between CoT+IFDNS and CoT+LoT across six logical reasoning datasets.
}
\centering
\scriptsize
\sisetup{table-format=2.2, 
table-number-alignment=center}
\begin{tabularx}{\textwidth}{c *{6}{X}}
\toprule

\multirow{2}{*}{\textbf{Methods}} &
\multicolumn{6}{c}{\textbf{Datasets}} \\
\cmidrule(lr){2-7}
& {\text{LogiQA}} & {\text{ReClor}} & {\text{RuleTaker}} & {\text{ProofWriter}} & {\text{PrOntoQA}} & {\text{FOLIO}} \\
\rowcolor{gray!20} 
\midrule
\multicolumn{7}{c}{\textbf{GPT-4o-mini}} \\
LoT+CoT       & 61.70 & 72.50 & 63.80 & 70.30 & 75.10 & 87.56 \\
IFDNS+CoT       & \textbf{63.70} & \textbf{77.00} & \textbf{68.50} & \textbf{74.90} & \textbf{76.80} & \textbf{89.63} \\
\midrule
\rowcolor{gray!20} 
\multicolumn{7}{c}{\textbf{GPT-4o}} \\
LoT+CoT       & 67.30 & 87.10 & 72.20 & 78.90 & 86.20 & 83.85 \\
IFDNS+CoT       & \textbf{75.10} & \textbf{89.30} & \textbf{79.00} & \textbf{83.00} & \textbf{92.70 }& \textbf{91.85} \\
\midrule
\rowcolor{gray!20} 
\multicolumn{7}{c}{\textbf{DeepSeek-V3}} \\
LoT+CoT       &73.20	&85.00	&72.60	&81.10	&89.50	&81.48\\
IFDNS+CoT      &\textbf{77.10 }	&\textbf{87.60} 	&\textbf{78.00} 	&\textbf{85.80} 	&\textbf{93.50} 	&\textbf{85.93}\\
\bottomrule
\end{tabularx}

\label{tab:compare_lot}
\end{table}
 
\subsection{Comparative Study of CoT+IFDNS vs and CoT+LoT}
\textbf{Comparative Analysis.} As shown in Table~\ref{tab:compare_lot}, we extend the main experiments by incorporating evaluations of the \texttt{DeepSeek-V3} model across six reasoning datasets, enabling comprehensive comparison and in-depth analysis of the differences between CoT+LoT and CoT+IFDNS methods. Our observations reveal that across all 18 experimental configurations ($3 ~\text{models} \times 6 ~\text{datasets}$), CoT+IFDNS demonstrates consistent improvements over CoT+LoT. Notably on the challenging RuleTaker dataset with state-of-the-art LLMs: \texttt{GPT-4o}: +6.8\% improvement,\texttt{DeepSeek-V3}: +5.4\% improvement. 

\textbf{Mechanism Analysis.} IFDNS implements multi-round feedback operations during both causal statement extraction and proposition/implication recognition phases, ensuring completeness and accuracy in symbolic information extraction. Subsequent application of generalized logical deduction rules ultimately enhances LLMs' logical reasoning capabilities. Detailed case studies are provided in Section~\ref{sec:case}.

\subsection{Comparative Study of CoT+IFDNS vs and Logic-LM}
\begin{figure}[t]
    \centering
    \begin{subfigure}{0.45\textwidth}
        \includegraphics[width=\textwidth]{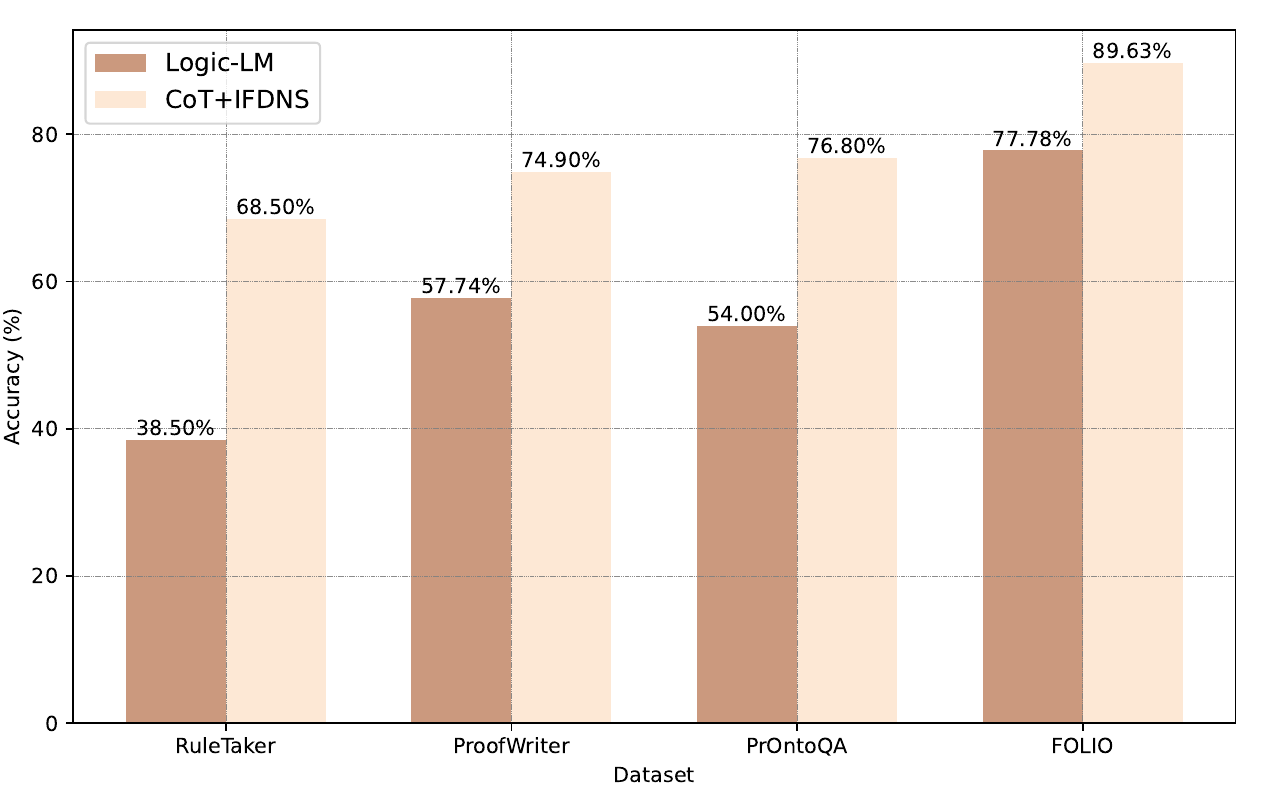} 
        \caption{Comparison between CoT+IFDNS and Logic-LM on \texttt{GPT-4o-mini}.}
        \label{fig:figure3}
    \end{subfigure}
    \hfill 
    \begin{subfigure}{0.45\textwidth}
        \includegraphics[width=\textwidth]{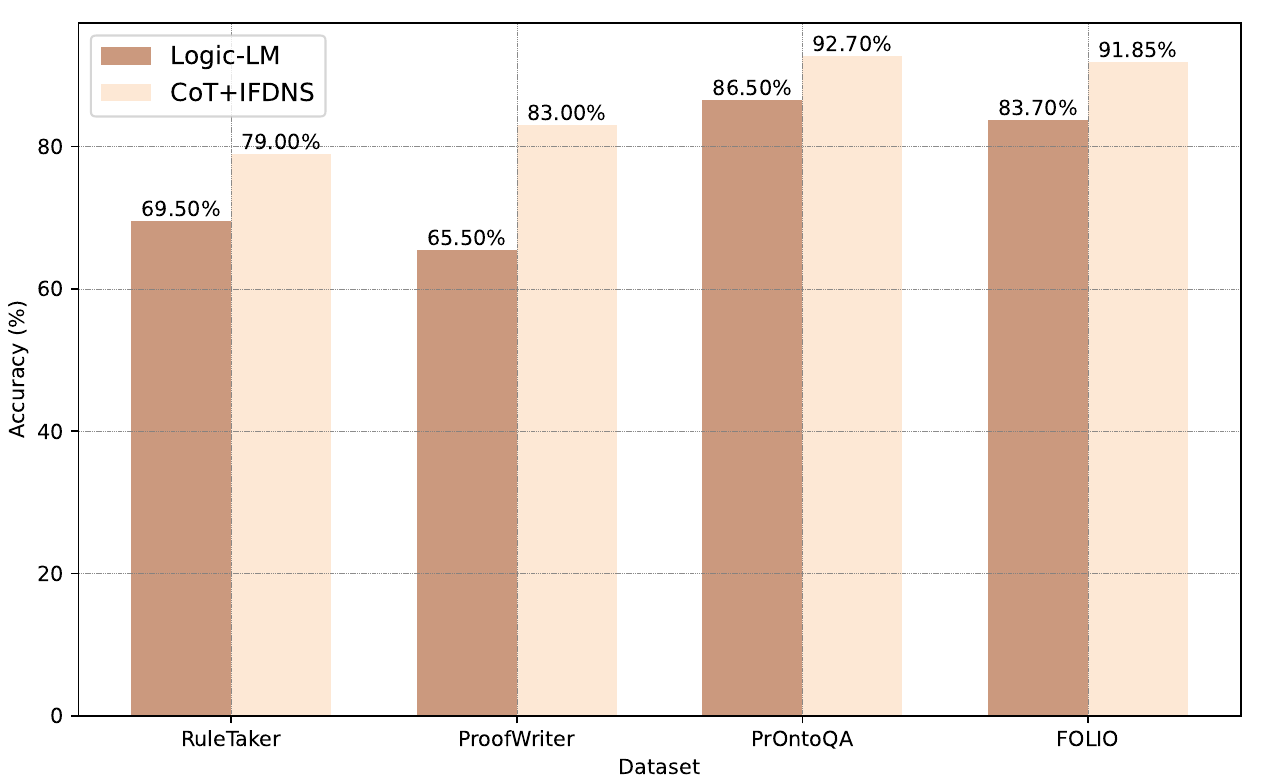} 
        \caption{Comparison between CoT+IFDNS and Logic-LM on \texttt{GPT-4o}.}
        \label{fig:figure4}
    \end{subfigure}
    \caption{Comparison between CoT+IFDNS and Logic-LM} 
    \label{fig:combined}
\end{figure}

As shown in Figure \ref{fig:combined}, we conducted a comparative analysis of the CoT+IFDNS method and the Logic-LM method on two LLMs and four datasets. Since the Logic-LM method calls the corresponding external symbol parser for specific datasets (e.g., ProntoQA, ProofWriter, and RuleTaker datasets use Pyke, while FOLIO uses Prover9), the LogiQA and ReClor datasets are not applicable to the Logic-LM method. The experimental results reveal that the performance of CoT+IFDNS on four datasets has improved by 30.00\%, 17.16\%, 22.80\%, and 11.85\% under the \texttt{GPT-4o-mini} (see Figure \ref{fig:figure3}). In comparison, under the \texttt{GPT-4o}, the performance improvements are 9.50\%, 17.50\%, 6.20\% and 8.15\% (see Figure \ref{fig:figure4}), indicating that the enhancements achieved with the \texttt{GPT-4o-mini} are significantly greater than those with the \texttt{GPT-4o}. It is worth noting that the effect of the Logic-LM method is even lower than that of the ordinary CoT prompt method in some cases. We speculate that this situation may be due to the complexity of the Logic-LM method. Specifically, the Logic-LM method first converts different datasets into symbolic forms such as logic programming language (LP) and first-order logic (FOL) through LLM, and then calls different symbol solvers to derive the final answer. However, the more complex the generated symbolic form is, the higher the performance requirements for the LLM are, and its effect is highly dependent on the success rate of the symbol solver. During the experiment, we found that when generating LP or FOL using \texttt{GPT-4o} and \texttt{GPT-4o-mini}, there are numerous syntax errors, causing the symbol solver to be unable to parse, especially more significant on the \texttt{GPT-4o-mini} with fewer parameters. In contrast, the IFDNS method only converts the original input into simple propositional logic and introduces a stepwise multi-round feedback module. Moreover, the IFDNS method will translate the extended logical formulas into natural language and re-add them to the original input to guide the large model to think step by step from scratch. This method largely avoids the problem of information loss that may occur when extracting symbols by LLM.

\begin{table}[t]
\caption{Three ablation variants' studies: comparison of accuracies of IFDNS+CoT and three ablation variants on four datasets, $\uparrow$ indicates the ablation effectiveness of IFDNS+CoT, bold numbers represent the best performance on each dataset. The experiments were conducted under the \texttt{GPT-4o}.}
\centering
\scriptsize
\sisetup{table-format=2.2, table-number-alignment=center}
\begin{tabularx}{\textwidth}{c *{4}{X}}
\toprule
\multirow{2}{*}{\textbf{Ablation variants}} & 
\multicolumn{4}{c}{\textbf{Datasets}} \\
\cmidrule(lr){2-5}
& {\text{Ruletaker}}$\uparrow$ & {\text{PrOntoQA}}$\uparrow$ & {\text{R-GSM}}$\uparrow$ & {\text{LogiQA}}$\uparrow$ \\
\midrule
w/o Multi-round Feedback        & 71.40 & 89.50 & 88.73 & 70.80 \\
w/o Subject/Quantifier Enhancement      & 72.20 & 90.30 & 87.27 & 72.20 \\
w/o Reorder      & 71.40 & 87.50 & 87.73 & 70.60 \\
\midrule
CoT+IFDNS                    & \textbf{79.00} & \textbf{92.70} & \textbf{90.73} & \textbf{75.10} \\
\bottomrule
\end{tabularx}

\label{tab:ablation_study}
\end{table}

\subsection{Ablation Experiment Results}\label{sec:ablation} 
As shown in Table \ref{tab:ablation_study}, our systematic module ablation study reveals the contribution of each component of the framework to the final performance. On the RuleTaker dataset, the complete CoT+IFDNS method (79.00\%) achieves absolute improvements of 7.6\%, 6.8\%, and 7.6\% compared to the variants without multi-round feedback, without subject/quantifier enhancement, and without reordering, respectively. This verifies the critical role of multi-round feedback in ensuring the quality of extracted implication expressions. In particular, on the PrOntoQA dataset, which requires precise subject distinction, the variant lacking subject/quantifier enhancement (90.30\%) shows a 2.4\% gap compared to the complete method (92.70\%), indicating that fine-grained recognition of proposition subjects effectively avoids symbol confusion. The reordering module exhibits the most significant impact in multi-hop reasoning scenarios: on the PrOntoQA dataset, removing this module leads to a 5.2\% decrease in accuracy, directly related to the dataset’s random ontology traversal direction. In the R-GSM dataset, where the key order of statements is disrupted, the performance degradation caused by the absence of reordering (90.73\%$\rightarrow$87.73\%) further demonstrates the role of premise order optimization in promoting complex reasoning. The experimental results show that the three core components—multi-round feedback, subject/quantifier enhancement, and reordering work together synergistically across different dimensions to build a complete technical path for enhancing the logical reasoning ability of LLMs.

\subsection{Case Study}
\label{sec:case} 
\begin{figure}[t]
    \centering
    \includegraphics[width=1\linewidth]{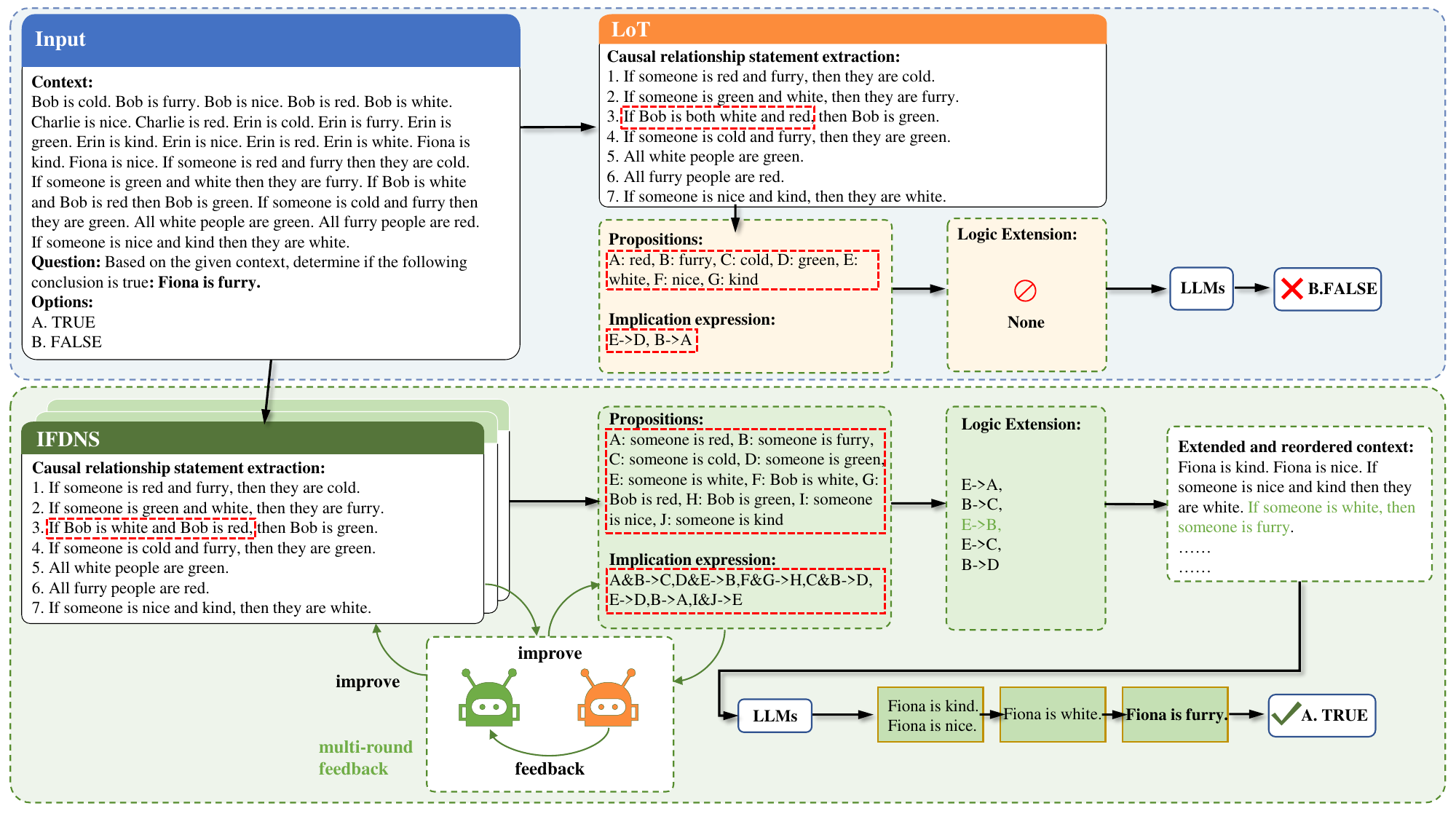}
    \caption{Comparison case study between IFDNS and LoT from the ProofWriter dataset. The red dashed rectangles highlight the differences between the two methods. LoT lacks the ability to understand all the propositions contained in the context and extract corresponding implication expressions, thus hindering logical extension. In contrast, IFDNS performs multi-round feedback on each stage of causal statement extraction, proposition and implication expression extraction. It utilizes broader logical deduction laws to extract implicit relationships and seamlessly integrates the additional information back into the original prompt, ultimately enhancing the LLM’s ability to generate accurate results.}
    \label{fig:case_study}
\end{figure}

Based on our experiments, we have conducted a comparative case study of LoT and IFDNS methods, as shown in  Figure \ref{fig:case_study}. In the process of extending the original context, LoT will cause information loss and errors. Specifically, when LoT extracts causal relationship statements with relational meanings from the original context, such as "If Bob is both white and red, then Bob is green.", it exceeds the upper limit of LoT's ability to recognize propositions and logic, making it impossible for LoT to extract useful information from this sentence. LoT further obtains propositions such as "A: red, B: furry, C: cold, D: green, E:white, F: nice, G: kind" from causal relationship statements, but the propositions obtained are too general. The original context "Bob is white" and "All white people" clearly correspond to two subjects, but should be extracted as two propositions. However, LoT can only extract "E:white", ultimately causing information loss and inability to further extend the extracted expressions. In contrast, IFDNS successfully extracts accurate causal relationship statements and propositions, which is attributed to the step-by-step verification mechanism of IFDNS. IFDNS extracts different propositions for different subjects, demonstrating its ability to identify different subjects with fine granularity. In the logical extension stage, IFDNS uses relevant laws of propositional logic to perform dynamic logical extension, derives multi-layer implicit relationships through Python, and finally translates these implicit relationships into natural language and adds them to the original context, enabling the LLMs to directly derive complex multi-hop problems in sequence, thereby enhancing the logical reasoning ability of the LLMs.

\section{Related Work}
\subsection{Prompt-based LLM Reasoning}
Logical reasoning, the process of deriving correct conclusions from context, premises, and questions, is significantly enhanced by prompt-based learning \cite{liu2023pre}, which provides crucial insights for improving the logical reasoning capabilities of LLMs. CoT \cite{cot} employs multiple few-shot to simulate the step-by-step reasoning process of humans, significantly boosting the performance of LLMs in arithmetic, symbolic and commonsense reasoning tasks. However, its effectiveness heavily depends on model scale and the quality of human-designed prompts. To overcome these limitations, subsequent research, such as Zero-shot CoT \cite{zero-shot}, combines zero-shot learning with CoT, directly activating reasoning capabilities through a single prompt. CoT-SC \cite{con-sc} selects the optimal answer through multi-path sampling and majority voting but faces high computational costs. ToT \cite{tree}, GoT \cite{graph}, and CR \cite{cumulative} all support non-linear reasoning, combining state evaluation and search algorithms to aggregate multiple thought chains, thereby enhancing complex problem-solving abilities. Least-to-Most \cite{zhou2022least} decomposes complex problems into a sequence of subproblems, reducing reasoning difficulty through phased solutions, though its generalization is limited by prompt design. Additionally, CauseJudger \cite{he2024causejudger} employs abductive logical reasoning by converting thought chains from reverse to forward and removing irrelevant information to identify the authenticity of potential causes. 

Although these methods enhance reasoning capabilities of LLMs, they still suffer from the accumulation of errors in intermediate steps and issues of hallucination.

\subsection{Neuro-symbolic Approaches for Logical Reasoning}
To address the shortcomings of LLMs in symbolic logical reasoning, researchers have proposed neuro-symbolic methods that integrate LLMs with symbolic reasoning. Faithful Chain-of-Thought \cite{faithful} ensures the faithfulness of reasoning through formal logical constraints. PAL \cite{pal} and PoT \cite{pot} leverage LLMs to read natural language inputs, generate Python code, and execute it via an interpreter. Logic-LM \cite{panlogiclm23} decomposes logical reasoning problems into three stages: problem formulation, symbolic reasoning, and result interpretation, ensuring transparency and faithfulness in the reasoning process via a symbolic solver. Building on this direction, SatLM \cite{ye2024satlm} and LINC \cite{oglz_linc_2023} convert natural language reasoning into satisfiability problems and first-order logic, respectively, and solve them using theorem provers. LoT \cite{liu2024logic} leverages propositional logic to generate extended logical information from input contexts, enhancing reasoning capabilities by supplementing the input prompt with this additional logical information. However, existing methods commonly face information loss: deviations in mapping natural language to formal logic during symbolization may introduce errors in intermediate reasoning steps. Additionally, premise order sensitivity \cite{prontoqa,premise_order} reveals that LLMs’ reliance on the ordering of premises further exacerbates instability in reasoning outcomes.

\subsection{Verification and Utilization of Error Information in the Logical Reasoning Process of LLM}
LLMs are prone to hallucinations in complex reasoning tasks. To address this, researchers have proposed methods like Self-Refine \cite{madaan2024self} and Self-Verification \cite{self-verification}, which refine initial outputs through iterative feedback. However, these approaches optimize answers by globally evaluating reasoning chains, often failing to detect errors in intermediate steps. In contrast, Self-Check \cite{miao2023selfcheck} and Deductive Verification \cite{ling2024deductive} progressively verify reasoning steps, while MAD \cite{liang2023encouraging} and JoT \cite{park2024judgment} employ multi-agent role-playing to iteratively validate results. CoT-Rerailer \cite{cot-rerailer} combines multi-agent debate with step-level evaluators for error localization and path correction. XoT \cite{xot} integrates three reasoning paradigms: Program-of-Thought (PoT) \cite{pot}, Equation-of-Thought (EoT) \cite{xot}, and CoT \cite{cot}. Building on these, Wrong-of-Thought (WoT) \cite{zhang2024wrong} introduces multi-perspective validation and incorporates error information into prompts to avoid recurring mistakes.

\section{Conclusion}
In this work, we propose a novel neuro-symbolic framework IFDNS, aimed at addressing the inherent symbol extraction information loss challenge present in existing neuro-symbolic methods. 
By implementing multi-round feedback during the stages of causal statement extraction and proposition/implication expression extraction respectively, IFDNS effectively reduces errors caused by intermediate hallucinations and premise order sensitivity. This effectively mitigates information loss issues and enhances the logical reasoning capabilities of LLMs. We conducted extensive validation on six challenging logical reasoning datasets (ReClor, LogiQA, RuleTaker, ProofWriter, FOLIO, and PrOntoQA). The results demonstrate that IFDNS can significantly improve the performance of existing prompt methods such as CoT and CoT-SC across a wide range.

\section{Limitations}
Although our proposed method supports some laws of propositional logic, its ability to perform logical induction is actually limited and its complexity is not as high as that of higher-order logic. In the future, we will attempt to continue adding some additional connectives and logical reasoning laws or introduce other logical systems to further enhance its logical expression ability and thereby improve its logical reasoning ability. Meanwhile, the method in this paper adopts a step-by-step verification approach, that is, a multi-round feedback method in the sub-stage of extracting neural symbols. This will to some extent increase the consumption of tokens. In the future, we will strive to reduce token consumption and further promote this method. Additionally, there may be other better forms of using the method of resetting the word order of the final context to achieve a better effect of ignoring the word order of the input prompt.


%
%
%
\bibliographystyle{splncs04}
\bibliography{reference}

\end{document}